\documentclass[a4paper,fleqn，]{cas-sc}

\usepackage[authoryear]{natbib}
\usepackage{algorithmic}
\usepackage{algorithm}

\def\tsc#1{\csdef{#1}{\textsc{\lowercase{#1}}\xspace}}
\tsc{WGM}
\tsc{QE}
\tsc{EP}
\tsc{PMS}
\tsc{BEC}
\tsc{DE}

\begin{document}
\let\WriteBookmarks\relax
\def\floatpagepagefraction{1}
\def\textpagefraction{.001}
\shorttitle{CaRe-Ego}
\shortauthors{Y. Su et~al.}

\title [mode = title]{CaRe-Ego: \underline{C}ontact-\underline{a}ware \underline{Re}lationship Modeling for \underline{Ego}centric Interactive Hand-object Segmentation}   


\author{Yuejiao Su}[                        orcid=0009-0006-9118-9217]
\ead{yuejiao.su@connect.polyu.hk}

\credit{Conceptualization of this study, Methodology, Software, Writing}

\affiliation{organization={
                Department of Electrical and Electronic Engineering, The Hong Kong Polytechnic University},
            addressline={Hung Hum},
           city={Kowloon},
           country={Hong Kong SAR}}

\author{Yi Wang}[                        orcid=0000-0001-8659-4724]
\ead{yi-eie.wang@polyu.edu.hk}
\credit{Conceptualization of this study, Methodology, Supervision, Revision}

\author{Lap-Pui Chau}[                        orcid=0000-0003-4932-0593]
\cormark[1]
\ead{lap-pui.chau@polyu.edu.hk}

\credit{Conceptualization of this study, Methodology, Supervision, Revision}

\cortext[cor1]{Corresponding author}

\begin{abstract}
Egocentric Interactive hand-object segmentation (EgoIHOS) requires the segmentation of hands and interacting objects in egocentric images, which is crucial for understanding human behavior in assistive systems. 
Previous methods typically recognize hands and interacting objects as distinct semantic categories based solely on visual features, or simply use hand predictions as auxiliary cues for object segmentation. 
Despite the promising progress achieved by these methods, they fail to adequately model the interactive relationships between hands and objects while ignoring the coupled physical relationships among object categories, ultimately constraining their segmentation performance.
To make up for the shortcomings of existing methods, we propose a novel method called CaRe-Ego that achieves state-of-the-art performance by emphasizing the contact between hands and objects from two aspects. 
First, we introduce a Hand-guided Object Feature Enhancer (HOFE) to establish the hand-object interactive relationships to extract more contact-relevant and discriminative object features.
Second, we design the Contact-centric Object Decoupling Strategy (CODS) to explicitly model and disentangle coupling relationships among object categories, thereby emphasizing contact-aware feature learning.
Experiments on various in-domain and out-of-domain test sets show that Care-Ego significantly outperforms existing methods with robust generalization capability.
Codes are publicly available at {\url{https://github.com/yuggiehk/CaRe-Ego/}}.
\end{abstract}




\maketitle

\section{Introduction}
\label{sec:intro}

Recent advancements in edge computing and computer vision \citep{ZHANG2025126852,TRAVER2022116079,XU2025126962,LUO2025125431} have engendered a growing interest in Head-Mounted Devices (HMD) such as the Apple Vision Pro \citep{DBLP:journals/corr/abs-2401-08685} and Microsoft Hololens 2 \citep{DBLP:conf/vr/ZhangHKY23}.
Consequently, there is a rising trend in capturing abundant egocentric (or first-person view, FPV) images and videos.
Compared with the third-person view (TPV) or exocentric data \citep{DBLP:journals/tmm/LiCWX24}, egocentric data reflects visual signals from the perspective of the camera wearer, emphasizing individual behavior across diverse contexts with noticeable background changes \citep{su2025annexeunifiedanalyzinganswering}.
This perspective allows for a more comprehensive exploration of questions related to individual activities, such as ``What is this person doing?''
When analyzing egocentric visual data, critical cues for interpretation emerge from the interactions between the wearer's hands and the objects they contact.

\begin{figure}
\centering
\includegraphics[width=\textwidth]{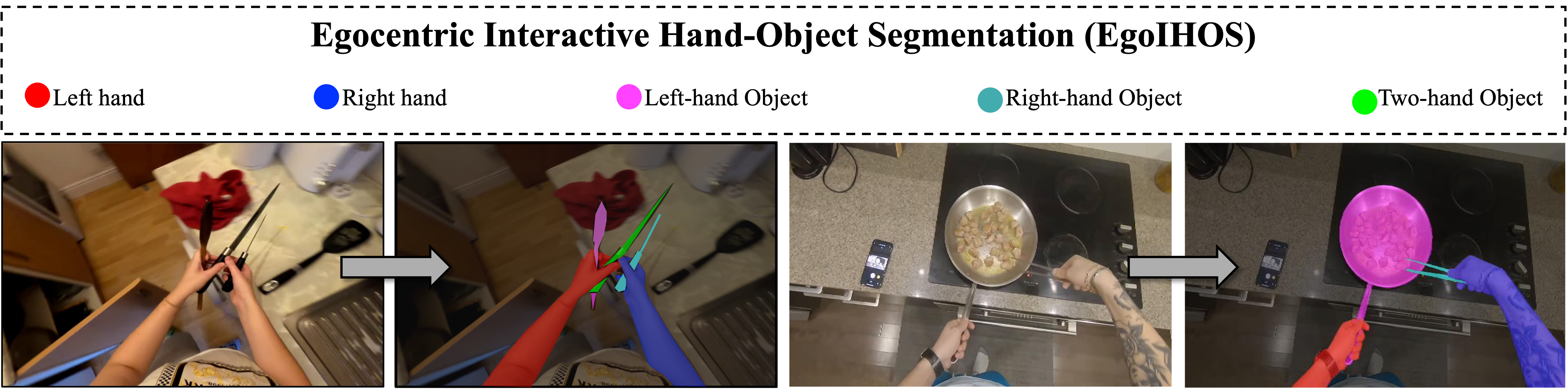}
\caption{\textbf{Illustration of EgoIHOS task}. This task aims to segment the input egocentric image into five categories: left hand, right hand, left-hand objects, right-hand objects, and two-hand objects.
}
\label{fig_1}
\end{figure}

Therefore, to delve deeply into human behavior encoded within egocentric images at a fine-grained level, Egocentric Interactive Hand-Object Segmentation (EgoIHOS) \citep{DBLP:conf/eccv/ZhangZSS22} is proposed aiming to perform pixel-level segmentation of \emph{hands} and \emph{objects that are interacting with hands} in egocentric images, as shown in Fig. \ref{fig_1}.
This unexplored segmentation task enables a precise response to the query: ``Where are the left and right hands, and what objects are they interacting with?'' which can function as an essential algorithm serving as the foundation for various subsequent applications, such as augmented reality/virtual reality (AR/VR) \citep{DBLP:journals/vr/LinLW23,DBLP:journals/pacmhci/ShiWQHL23}, embodied AI \citep{DBLP:journals/corr/abs-2403-16182,DBLP:journals/corr/abs-2311-18259,DBLP:conf/nips/ChangP023,zheng2024ocrm}, and medical assistive systems for visually impaired individuals \citep{DBLP:conf/nips/LiWZULYNPG23,DBLP:conf/miccai/AzadAAKM23}.

Current EgoIHOS approaches can be categorized into two distinct pipelines: one-stage and multi-stage methods.
One-stage methods \citep{DBLP:conf/eccv/ZhangZSS22,DBLP:conf/nips/XieWYAAL21,DBLP:conf/cvpr/Jain0C0OS23,DBLP:conf/eccv/CaoWCJZTW22,DBLP:conf/eccv/XiaoLZJS18} typically consider the EgoIHOS task as a conventional visual segmentation problem, \emph{i.e.}, regard hands and interacting objects as different semantic categories and recognize them based solely on extracted visual features.
However, since the EgoIHOS task aims to segment \emph{objects interacting with hands} rather than fixed categories, one-stage methods ignore the interactive factors between hands and objects, potentially leading to a significant decrease in segmentation performance. 
Additionally, regardless of variations in interactions and environments, identifying hands in egocentric images is often more straightforward due to their recognizable shape and texture.
To address the limitations of one-stage methods, multi-stage methods \citep{DBLP:conf/eccv/ZhangZSS22} are proposed to predict the mask of the hand as an auxiliary signal to assist in the sequential interacting object segmentation.
 While multi-stage methods establish a coarse association between hands and objects, issues such as error accumulation and inadequate modeling of interactive relationships still cause negative impacts on accuracy.

It is important to highlight that the EgoIHOS task involves predicting three categories of interacting objects: \emph{left-hand objects, right-hand objects}, and \emph{two-hand objects}, as depicted in Sec. \ref{sec::task}. 
We observe that the \emph{two-hand object} inherently contains attributes from both \emph{left-hand object} and \emph{right-hand object} because it represents a single object simultaneously interacting with both hands.
However, existing one-stage and multi-stage approaches \citep{DBLP:conf/eccv/ZhangZSS22,DBLP:conf/nips/XieWYAAL21,DBLP:conf/cvpr/Jain0C0OS23,DBLP:conf/eccv/CaoWCJZTW22,DBLP:conf/eccv/XiaoLZJS18,DBLP:conf/eccv/ZhangZSS22} commonly neglect these coupled relationships and process different categories of interacting objects as discrete entities.
As a result, the network is trained to: 1) segment all objects contacted by left and right hands independently, and 2) classify the segmented objects based on ``whether the object is interacting with both hands simultaneously'' to distinguish the \emph{two-hand object}.
We argue that the second step - classification - not only introduces an additional training burden but also shifts the model's focus from learning interaction and contact, potentially degrading overall precision. Furthermore, this classification step becomes redundant when effectively modeling the inherent relationships among the three categories of interacting objects.

To address the shortcomings of the existing methods, this paper proposes a novel \textbf{C}ontact-\textbf{a}ware \textbf{Re}lationship Modeling Network for \textbf{Ego}IHOS (\textbf{CaRe-Ego}) to promote the model concentrate more on the interaction between hands and objects from two aspects.
First, to establish the hand-object correlation explicitly, we design the Hand-guided Object Feature Enhancer (HOFE) that leverages hand features as prior knowledge to extract contact-relevant and highly discriminative object representations by the hand-guided cross-attention mechanism.
Second, we present a Contact-centric Object Decoupling Strategy (CODS) to disentangle the relationships among diverse object categories. Through the CODS, the model is restricted to segmenting all objects interacted by left and right hands without classifying ``whether the object is touched by two hands simultaneously,'' emphasizing hand-object contact and significantly reducing category confusion.
By integrating HOFE and CODS, our CaRe-Ego framework effectively benefits from relationship modeling between \emph{hands and objects} as well as \emph{objects and objects}, emphasizing hand-object contact learning and significantly improving segmentation accuracy. 
The main contributions of this work can be summarized as follows:

\begin{itemize}
\item To effectively address the EgoIHOS task, we propose the CaRe-Ego framework, which focuses on interactive feature learning through modeling the relationships between \emph{hands and objects}, as well as \emph{objects and objects}.
   \item 
To explicitly model hand-object interactions in EgoIHOS, we introduce the HOFE. This module utilizes hand-guided cross-attention to leverage hand features as prior knowledge, facilitating the extraction of contact-relevant and representative object representations. 
    \item 
    Inspired by our observations of interacting object categories, we present the innovative CODS. This approach effectively models and disentangles the coupled relationships among diverse object categories, reducing classification confusion while emphasizing the learning of hand-object contacts.
    \item Comprehensive experiments are conducted on various in-domain and out-of-domain test sets to evaluate the performance and generalization ability of the proposed CaRe-Ego. The experimental results demonstrate the superiority and robust generalization ability of our method over existing state-of-the-art methods.
\end{itemize}

\section{Related Work}
This section mainly introduces the related work of this paper. Since the EgoIHOS task is part of Egocentric Hand-Object Interaction (EgoHOI), we will first introduce EgoHOI (Sec. \ref{related::egohoi}) and then refine to EgoIHOS (Sec. \ref{related:egohos}).

\subsection{Egocentric Hand-Object Interaction}
\label{related::egohoi}
Egocentric images and videos offer a clear window into how humans physically engage with their surroundings and objects using their hands. In recent years, the interpretation and analysis of egocentric visual data have earned increasing attention from the research community, especially in the field of Egocentric hand-object interaction (EgoHOI) \citep{chatterjee2024opening,DBLP:conf/cvpr/GraumanWBCFGH0L22,mangalam2024egoschema} due to the release of various large-scale datasets such as Ego4D, EPIC-KITCHENS, HOI4D, etc. \citep{DBLP:journals/corr/abs-1804-02748,DBLP:conf/cvpr/GraumanWBCFGH0L22,goyal2017something,sigurdsson2018charades,xu2023egopca}.
EgoHOI aims to predict the bounding boxes or masks for hands and interacting objects, which is quite challenging due to the background change and severe occlusion.

Although large-scale egocentric datasets have recently emerged, the available FPV data remains substantially more limited than TPV datasets. Several studies \citep{li2021ego,xu2023pov} have explored cross-view representation learning to address this data limitation, aiming to transfer view-agnostic knowledge from TPV to FPV domains. However, these approaches typically require precisely aligned multi-view visual data, which is labor-intensive and time-consuming.
Complementary to these approaches, several methods \citep{huang2018predicting,li2018eye,lin2022egocentric,pramanick2023egovlpv2} have incorporated multi-modal cues (e.g., gaze and textual context) to enrich the learned representations. Despite significant advances through the latest transformer-based architectures \citep{zhang2023helping,shiota2024egocentric,9591443,liu2022joint}, current EgoHOI methods remain limited by lacking explicit interactive relationship modeling for hands and objects.

\subsection{Egocentric Hand-Object Interactive Segmentation}
\label{related:egohos}

Extended from EgoHOI, EgoIHOS \citep{DBLP:conf/eccv/ZhangZSS22} is an unexplored problem aiming to perform fine-grained pixel-level segmentation for hands and interacting objects.
The final prediction output encompasses a set of categories, including left-hand, right-hand, left-hand objects (objects interacting with left hand only), right-hand objects (objects interacting with right hand only), and two-hand objects (objects simultaneously interacting with both hands).
Current EgoIHOS methods can be divided into two distinct pipelines: one-stage and multi-stage methods.

One-stage methods \citep{DBLP:conf/eccv/ZhangZSS22,DBLP:conf/nips/XieWYAAL21,DBLP:conf/cvpr/Jain0C0OS23,DBLP:conf/eccv/CaoWCJZTW22,DBLP:conf/eccv/XiaoLZJS18,10655311} typically treated the EgoIHOS as the conventional visual segmentation task, which considered hands and interacting objects as distinct semantic categories, obtaining all predictions after feature extraction and mask generation. 
Although these methods have completed the EgoIHOS task end-to-end, they ignored the interactive factors between hands and objects, which may result in limited segmentation performance.
To overcome the shortcomings of one-stage methods, multi-stage approaches \citep{DBLP:conf/eccv/ZhangZSS22,zhang2023helping} were proposed to utilize the hand predictions as auxiliary cues to assist the sequential interacting object segmentation step-by-step. 
But they still suffered from error accumulation and inadequate interactive relationship modeling, resulting in suboptimal performance.
Furthermore, both one-stage and multi-stage methods regarded the different categories of interacting objects as independent and ignored the inherent coupling relationships between them, causing unnecessary confusion in classifying the object categories during training.
Unlike the previous methods, the proposed CaRe-Ego explicitly models the interactive relationships between hand and object features to extract contact-aware object features through HOFE. The CODS is also incorporated in CaRe-Ego to decouple the object correlations and emphasize the interaction, reducing the network's confusion about classification.

\section{Methodology}
This section introduces the proposed CaRe-Ego in detail from top to bottom. 
We begin by describing the EgoIHOS task in Sec. \ref{sec::task}. 
We then elaborate on the overall architecture of CaRe-Ego in Sec. \ref{sec::overview}.
The detailed methodologies of hand-guided object feature enhancer (HOFE) and contact-centric object decoupling strategy (CODS) are displayed in Sec. \ref{sec::HOR} and Sec. \ref{sec::ordm}, respectively.
Finally, we discuss the training and inference process for our method in Sec. \ref{sec::Training}.

\subsection{Task Description}
\label{sec::task}
  
This section delineates the detailed configuration of an unexplored task - EgoIHOS.
The EgoIHOS task aims to segment the hands and objects directly contacted by hands.
Giving an egocentric image $\textbf{I}\in \mathbb{R}^{H \times W \times 3}$, the goal of EgoIHOS is to predict the masks of the following entities: hands $\textbf{M}_{H}$ (including left and right hand), left-hand objects $\textbf{M}_{O}^{L}$ (objects directly interacting with the left hand), right-hand objects $\textbf{M}_{O}^{R}$ (objects directly interacting with the right hand), and two-hand objects $\textbf{M}_{O}^{T}$ (objects interacting with both hands simultaneously).

\begin{figure}
\centering
\includegraphics[width=\textwidth]{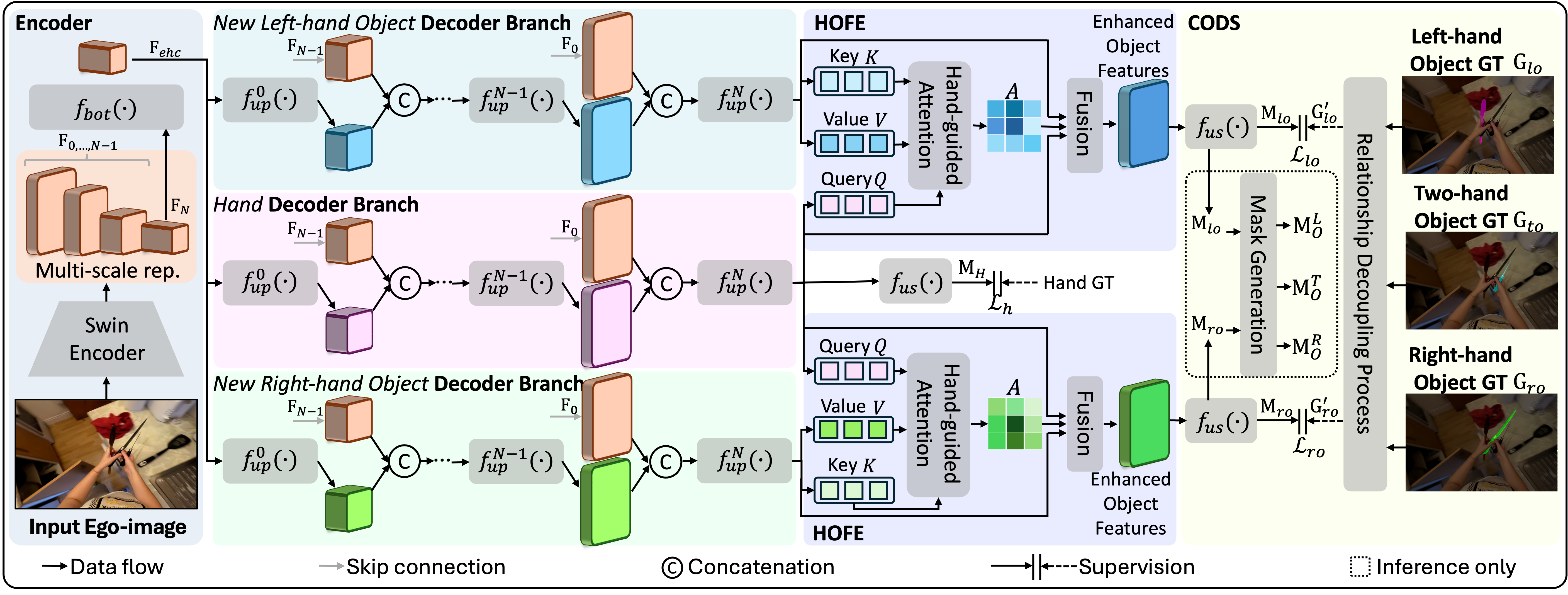}
\caption{\textbf{Overall diagram of the proposed CaRe-Ego.} The method comprises four main components: an encoder, a multi-branch decoder, a hand-guided object feature enhancer (HOFE) (Sec. \ref{sec::HOR}), and a contact-centric object decoupling strategy (CODS) (Sec. \ref{sec::ordm}). Rep. in this figure denotes representations. 
}
\label{FIG:1}
\end{figure}

\subsection{Overall Architecture}
\label{sec::overview}
The Overall diagram of the proposed CaRe-Ego is shown in Fig. \ref{FIG:1}.
The model consists of an encoder, a multi-branch decoder, the HOFE module, and the CODS.
Specifically, taking an egocentric image $\mathbf{I}$ as input, the first step of CaRe-Ego is to use the image encoder to extract the multi-scale representative features $\textbf{F} = \{\textbf{F}_i \in \mathbb{R}^{H_i \times W_i \times C_i} \; | \; i=0,1,\cdots,N\}$ by a series of down-samplings with a factor of 2, where $H_i $, $W_i$, and $C_i$ are the height, width, and channel of the $i^{th}$-stage feature map, respectively. The image encoder used in practice is the Swin Transformer encoder \citep{DBLP:conf/iccv/LiuL00W0LG21}, and N is set to 3. 
The extraction process of generating multi-scale features in the encoder can be expressed as follows:
\begin{gather}
    \mathbf{F}_{i+1} = f_{enc}^{i+1}(\mathbf{F_i} \; | \; W_{enc}^{i+1}) , \; i \in \left\{0,1,\cdots,N-1\right\},\\
    \mathbf{F_0} = f_{enc}^0(\mathbf{I} \; | \; W_{enc}^0),
\end{gather}
where the $f_{enc}^i(\cdot)$ denotes the $i^{th}$ encoder stage for generating the feature map $\mathbf{F}_{i}$, and the $W_{enc}^i$ represents corresponding parameters. 
After generating the multi-scale features, the final global feature map $\textbf{F}_{N}$ is sent into the bottleneck to enhance the representation of the encoded feature. Following the work \citep{DBLP:conf/eccv/CaoWCJZTW22}, the bottleneck is composed of a patch merging layer and two Swin Transformer blocks, which are denoted as follows:
\begin{equation}
    \mathbf{F}_{ehc} = f_{bot}(\mathbf{F}_{N} \; | \; W_{bot}),
\end{equation}
where the $\mathbf{F}_{ehc}$ represents the enhanced final encoded feature map, and the $f_{bot}(\cdot)$ denotes the bottleneck network with the parameter of $W_{bot}$.

The enhanced global feature $\mathbf{F}_{ehc}$ and multi-scale features $\textbf{F}^\prime=\{ \textbf{F}_i \; | \; i=0,1,\cdots,N-1\}$ are sent into decoder for mask prediction. 
As shown in Fig. \ref{FIG:1}, the decoder includes different branches for hands, new left-hand objects, and new right-hand objects. The explanation for new left-hand object and new right-hand object branches will be explained in the Sec. \ref{sec::ordm}. 
Following the previous work \citep{DBLP:conf/eccv/ZhangZSS22}, CaRe-Ego also contains the contact boundary (CB) decoder branch to predict the mask of the contact boundary. For simplification, we remove it in Fig. \ref{FIG:1}. 
In each decoder branch, N+1 decoder stages are utilized to recover the resolution and generate the masks. 
The output feature of each stage is concatenated with corresponding encoded features by skip-connection \citep{he2016deep,9428155} as the input to the next stage. The process of each decoder branch can be represented as follows,
\begin{gather}
    \textbf{D}_0 = f_{up}^0(\textbf{F}_{ehc}), \\
    \textbf{D}_{i+1} = f_{up}^{i+1}(f_{cat}(\textbf{D}_i,\textbf{F}_{N-1-i})), \; i=\{0,1,...,N\},
\end{gather}
where the $f_{up}^i(\cdot)$ means the $i^{th}$ decoder stage, which outputs $\textbf{D}_{i}$ as decoded feature map. In practice, we use the decoder stage in Swin-UNet \citep{cao2021swinunetunetlikepuretransformer} in each decoder branch. The $f_{cat}(\cdot)$ means the concatenation operation.

One of the most naive approaches is to directly obtain the final mask for each decoder branch. However, as stated in Sec. \ref{sec:intro}, most existing methods suffer from inadequately modeling the interactive relationships between hands and objects, which causes negative impacts on segmentation accuracy.
Thus, after $N+1$ stages of decoding, we design the hand-guided object feature enhancer (HOFE) to utilize the hand features as prior knowledge to extract more contact-relevant and representative object features. The detailed structure of HOFE will be described in Sec. \ref{sec::HOR}.  At the same time, the hand features are upsampled to obtain the predicted mask of hands (including left and right hands), which is supervised by the hand ground truth (GT).

Besides, based on the observed coupled relationships between diverse interacting object categories, we design the contact-centric object decoupling strategy (CODS) for dealing with three interacting objects. The description of CODS will be introduced in Sec. \ref{sec::ordm}.

\begin{figure}[!t]
\centering
\includegraphics[scale=0.6]{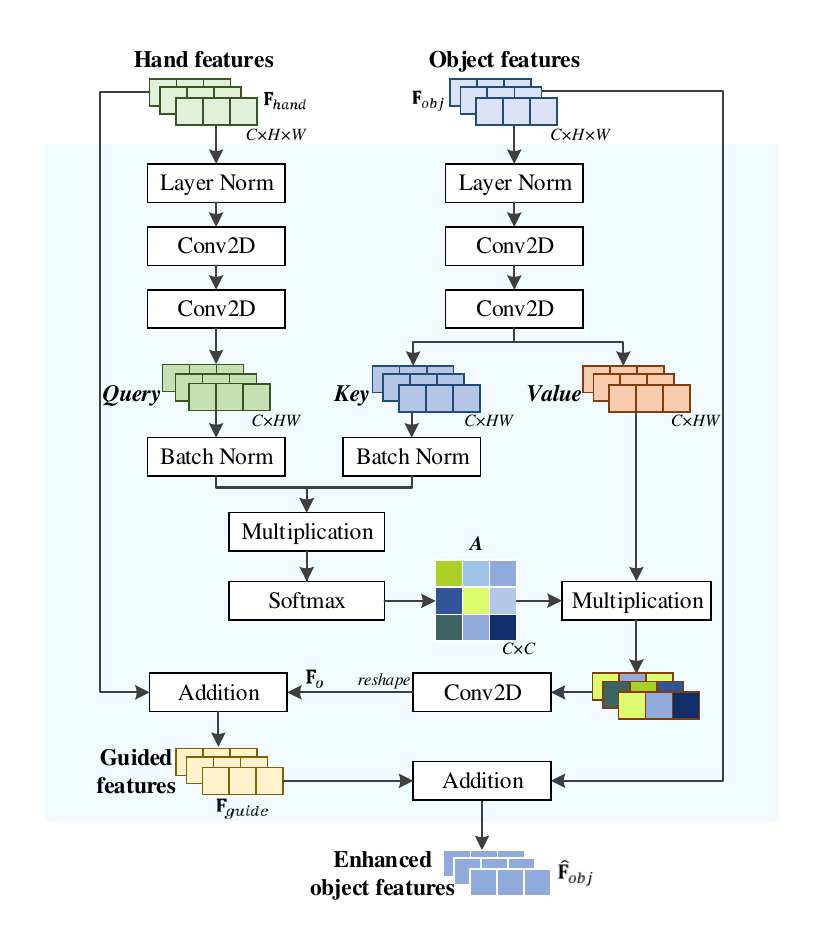}
\caption{\textbf{Detailed architecture of the proposed HOFE}. Taking the features of hands and objects as input, this module accomplishes hand-guided attention to model the relationship between the hands and interacting objects, promoting the contact-relevant object feature learning of the network.}
\label{fig_3}
\end{figure}

\subsection{Hand-guided Object Feature Enhancer}
\label{sec::HOR}
Among the egocentric visual information, the hand information of the HMD wearer is the most prominent and representative because we can easily recognize hands by shape and texture. 
However, we can only accurately identify interacting objects by whether they interact with hands, rather than relying solely on inherent attributes such as appearance features of objects, \emph{i.e.,} hands play a fundamental and decisive role in recognizing interacting objects.
So establishing the relationship between hands and interacting objects is necessary and beneficial to the EgoIHOS task. 
While certain sequential methods \citep{DBLP:conf/eccv/ZhangZSS22} employed predicted hand masks as auxiliary cues to segment interacting objects, they inadequately build an implicit connection between hands and objects, which causes suboptimal accuracy for segmentation. 

To explicitly establish the hand-object interactive relationship, we propose the hand-guided object feature enhancer (HOFE), which aims to use representative hand features as prior knowledge to guide the learning of contact-relevant object representations. 
As briefly shown in Fig. \ref{FIG:1}, the HOFE is employed twice to enhance the features of left-hand objects (upper) and right-hand objects (down), and the process in each HOFE is entirely consistent.
For simplicity, we will uniformly refer to both left-hand and right-hand objects as ``objects'' in this section.
The detailed structure of HOFE is displayed in Fig. \ref{fig_3}.

The input to HOFE includes two types of features: hand features $\textbf{D}_{hand} \in \mathbb{R}^{\hat{C} \times \hat{H} \times \hat{W}}$ and object features $\textbf{D}_{obj} \in \mathbb{R}^{\hat{C} \times \hat{H} \times \hat{W}}$ obtained from corresponding decoder branches, where $\hat{C}$, $\hat{H}$, and $\hat{W}$ are the channel, height, and width of the feature map.
The goal of the HOFE is to utilize the hand features as prior knowledge to help extract more contact-relevant and discriminative interacting object features.
Specifically, two stacked convolution operations are used after the LayerNorm layer to generate the query, key, and value first. 
The query $Q \in \mathbb{R}^{\hat{C} \times \hat{H}\hat{W}}$ is generated from hand features, and the key $K\in \mathbb{R}^{\hat{C} \times \hat{H} \hat{W}}$ and value $V\in \mathbb{R}^{\hat{C} \times \hat{H} \hat{W}}$ are generated by object features, which can be denoted as follows:
\begin{gather}
    Q = f_{conv}(f_{conv}(f_{ln}(\textbf{D}_{hand}))), \\
    K, V = f_{conv}(f_{conv}(f_{ln}(\textbf{D}_{obj}))),
\end{gather}
where the $f_{conv}(\cdot)$ and $f_{ln}(\cdot)$ represent the convolution and LayerNorm layers.
After that, the query, key, and value are used to perform the hand-guided attention operation, \emph{i.e.}:
\begin{gather}
\setlength{\abovedisplayskip}{1pt}
\setlength{\belowdisplayskip}{1pt}
    \textbf{A} = \delta (f_{Matmul}(Norm(Q), Norm(K^T))), \\
    \textbf{F}_o = f_{conv}(f_{Matmul}(\textbf{A}, V)),
\end{gather}
where the $\textbf{F}_o$ is the output after convolution, the $f_{Matnul}(\cdot)$ means the matrix multiplication, and the $\delta(\cdot)$ represents the softmax function. 
Then, in the fusion process, the input hand features are added to the $\textbf{F}_o$ to generate the guided features. Finally, the addition between guided features and original object features is performed to enhance the representation of the object features, which can be described as follows:
\begin{gather}
    \textbf{F}_{guide} = \textbf{F}_o + \textbf{D}_{hand},  \\
    {\textbf{D}}_{obj}^\prime = \textbf{D}_{obj} + \textbf{F}_{guide},
\label{equ8}
\end{gather}
where the ${\textbf{D}}_{obj}^\prime$ is the final enhanced object features.
In this way, hand features are used as queries to extract more representative and contact-relevant interacting object features. Therefore, the relationships between hands and interacting objects are established explicitly by the hand-guided attention mechanism, which can help the segmentation of interacting objects and improve the segmentation performance.

\subsection{Contact-centric Object Decoupling Strategy}
\label{sec::ordm}

One objective of the EgoIHOS task is to segment hand-interacting objects, which are categorized into three classes: left-hand objects, right-hand objects, and two-hand objects, as stated in Sec. \ref{sec::task}.
Current methods typically employ customized network architectures that are trained to predict these three object categories through supervised learning.
However, they ignored the inherent coupled relationships between these three object categories, \emph{i.e.,}  the two-hand object inherently contains attributes from both left-hand object and right-hand object because it represents a single object simultaneously interacting with both
hands. Assume a model is trained to predict left-hand objects, right-hand objects, and two-hand objects, it will go through two steps: 1) segment all
objects contacted by left- and right-hands independently, and 2) classify the segmented objects based on ``whether the
object is interacting with both hands simultaneously'' to distinguish \emph{two-hand objects}.
We argue that the second step - classification - not only introduces additional training confusion but also shifts the focus of the model from learning
interaction and contact, potentially degrading overall precision.

\begin{algorithm}[t]
\caption{CaRe-Ego}
\label{algorithm}
\begin{algorithmic}[1]
\REQUIRE Input egocentric image: $\textbf{I}$.
\REQUIRE Initialize: $W=\{W_{enc}, W_{dec}^{h}, W_{dec}^{lo}, W_{dec}^{ro}, W_{dec}^{cb}\}$.
\REQUIRE Hyper-parameters: $\alpha, \beta,\gamma, \lambda$.
\REQUIRE CODS - Relationship decoupling process: ${\textbf{G}}_{lo}^\prime = \textbf{G}_{lo} \bigcup \textbf{G}_{to},{\textbf{G}}_{ro}^\prime = \textbf{G}_{ro} \bigcup \textbf{G}_{to}$.
\STATE $\text{Model} = \text{CaRe-Ego}()$
\WHILE{training}
\STATE$ {\textbf{M}}_{lo},{\textbf{M}}_{ro},\textbf{M}_H, \textbf{M}_{CB} = \text{Model}(\textbf{I})$, predict all masks.
\STATE $\mathscr{L} =\alpha (l_{ce}(\textbf{M}_{lo},{\textbf{G}}_{lo}^\prime) + l_{ce}(\textbf{M}_{ro},{\textbf{G}}_{ro}^\prime)) + \gamma l_{ce}(\textbf{M}_H, {\textbf{G}}_h)+\lambda l_{ce}(\textbf{M}_{CB}, \textbf{G}_{cb}).$
\STATE $W \leftarrow W-\beta \nabla \mathscr{L}(W) $, Update Model
\ENDWHILE
\REQUIRE Input egocentric image for inference: $\textbf{I}_{inf}$.
\STATE $\text{Model}_{inf}$ = CaRe-Ego(W)
\WHILE{inference}
\STATE$ {\textbf{M}}_{lo},{\textbf{M}}_{ro},\textbf{M}_H, \textbf{M}_{CB} = \text{Model}_{inf}(\textbf{I}_{inf})$, predict all masks.
\STATE $\textbf{M}_O^T = {\textbf{M}}_{lo} \bigcap {\textbf{M}}_{ro}, \textbf{M}_O^L = {\textbf{M}}_{lo}\textbackslash \textbf{M}_O^T, \textbf{M}_O^R ={\textbf{M}}_{ro} \textbackslash \textbf{M}_O^T$, CODS - Mask generation.
\ENDWHILE
\STATE Return $\textbf{M}_H, \textbf{M}_{CB}, \textbf{M}_O^T, \textbf{M}_O^L,\textbf{M}_O^R$.
\end{algorithmic}
\end{algorithm}

To address these issues, we propose the CODS to establish the relations between object categories to eliminate the need for classification and emphasize the interactions during training.
Concretely, we introduce the concept of \emph{new left-hand objects} as all objects contacted by the left hand, which encompass the original left-hand objects and two-hand objects. 
 Similarly, \emph{new right-hand objects} include all objects that contact with the right hand, comprising original right-hand objects and two-hand objects.
Consequently, the two-hand objects are the \textbf{intersection} of the \emph{new left-hand objects} and \emph{new right-hand objects}.

During training, we use two decoder branches for \emph{new left-hand objects} and \emph{new right-hand objects}, entirely eliminating the two-hand objects. This approach encourages the model to focus exclusively on the interactions of the left and right hands without the need to classify whether an object is being interacted with by both hands simultaneously.
As shown in Fig. \ref{FIG:1}, after the HOFE outputs the enhanced object features for \emph{new left-hand objects} and \emph{new right-hand objects} using Equ. \ref{equ8}, the up-sampling layers $f_{us}(\cdot)$ are used to recover their resolution to obtain the predicted masks $\textbf{M}_{lo}$ and $\textbf{M}_{ro}$, respectively. 
Then the first step of CODS is to use the relationship decoupling process to generate supervision for these two branches based on the original left-hand object, right-hand object, and two-hand object GTs as follows,
\begin{gather}
    {\textbf{G}}_{lo}^\prime = \textbf{G}_{lo} \cup \textbf{G}_{to} = \{(x \in \textbf{G}_{lo}) \vee (x \in \textbf{G}_{to})\},  \\
    {\textbf{G}}_{ro}^\prime = \textbf{G}_{ro} \cup \textbf{G}_{to} = \{(x \in \textbf{G}_{ro}) \vee (x \in \textbf{G}_{to})\},
\end{gather}
where the $\cup$ means the union operation, the ${\textbf{G}}_{lo}$, ${\textbf{G}}_{ro}$, and ${\textbf{G}}_{to}$ are the original labels for left-hand objects, right-hand objects, and two-hand objects in the dataset. And the ${\textbf{G}}_{lo}^\prime$ and ${\textbf{G}}_{ro}^\prime$ represent the generated new supervision signals for new left-hand objects and new right-hand objects. Through this relationship decoupling process, the concept of two-hand objects is removed from network training.
${\textbf{G}}_{lo}^\prime$ and ${\textbf{G}}_{ro}^\prime$ are utilized to supervise \emph{new left-hand object} and \emph{new right-hand object} decoder branches, which can be expressed as follows,
\begin{gather}
\mathscr{L}_{lo} = l_{ce}(\textbf{M}_{lo}, \textbf{G}_{lo}^\prime), \\
\mathscr{L}_{ro} = l_{ce}(\textbf{M}_{ro}, \textbf{G}_{ro}^\prime),
\end{gather}
where the $\mathscr{L}_{lo}$ and $\mathscr{L}_{ro}$ denote the loss of new left-hand objects and new right-hand objects, which we use the cross entropy $l_{ce}(\cdot)$ in practice.

During inference, based on the predicted masks for \emph{new left-hand objects} and \emph{right-hand objects},
the CODS performs a mask generation process to obtain the final segmentation results required by the EgoIHOS, as shown in Fig. \ref{FIG:1}.
In specific, taking the $\textbf{M}_{lo}$ and $\textbf{M}_{ro}$ output from decoder branches, the two-hand object prediction
$\textbf{M}_{O}^T$ is generated by an ingenious operation - \textbf{intersection} between $\textbf{M}_{lo}$ and $\textbf{M}_{ro}$. Then the final predictions of the left-hand objects $\textbf{M}_{O}^L$ and right-hand objects $\textbf{M}_{O}^R$ have to remove the region of the two-hand object $\textbf{M}_{O}^T$. 
The mask generation process can be denoted as follows,
\begin{gather}
\label{equ:codsrefer}
    \textbf{M}_O^T = {\textbf{M}}_{lo} \cap {\textbf{M}}_{ro} =\{ (x\in\textbf{M}_{lo})\wedge (x \in \textbf{M}_{ro})\} , 
    \\
    {\textbf{M}}_O^L =  {\textbf{M}}_{lo} \textbackslash \textbf{M}_O^T =\{(x \in \textbf{M}_{lo}) \wedge (x \notin \textbf{M}_O^T)\}, \\ 
    {\textbf{M}}_O^R =  {\textbf{M}}_{ro} \textbackslash \textbf{M}_O^T=\{(x \in \textbf{M}_{ro}) \wedge (x \notin \textbf{M}_O^T)\},
    \label{equ:codsreferend}
\end{gather}
where the $\cap$ means the intersection operation and the set difference $A\textbackslash B$ contains elements in A that are not in B.
Through our proposed simple yet effective CODS, the two-hand objects category is completely decoupled during training, and the network is able to concentrate on the objects interacting with hands without classifying the categories, which decreases the training confusion and emphasizes the hand-object interactions. 
Notably, in inference, our proposed CODS is able to directly generate the masks required by EgoIHOS without additional training steps or optimization.

\subsection{Training and Inference}
\label{sec::Training}
The overall CaRe-Ego is trained by the weighted cross-entropy loss function between predictions and GTs of hands, new left-hand objects, new right-hand objects, and contact boundaries \citep{DBLP:conf/eccv/ZhangZSS22}. 
The overall loss function $\mathscr{L}$ can be computed by:
\begin{align} 
    \mathscr{L} =\alpha (l_{ce}(\textbf{M}_{lo},{\textbf{G}}_{lo}^\prime) + l_{ce}(\textbf{M}_{ro},{\textbf{G}}_{ro}^\prime)) + \\ \nonumber
    \gamma l_{ce}(\textbf{M}_H, {\textbf{G}}_h)+\lambda l_{ce}(\textbf{M}_{CB}, \textbf{G}_{cb}),
\end{align}
where $\alpha, \; \gamma,$ and $\lambda$ denote the loss weight of objects, hands, and contact boundaries, respectively. 
$\textbf{M}_{CB}$ presents the output of the contact boundary decoder branch, which is supervised by $\textbf{G}_{cb}$. Following previous work \citep{DBLP:conf/eccv/ZhangZSS22}, the GT of the contact boundary $\textbf{G}_{cb}$ is generated by the overlapping region of the dilated hand and object masks.
$\textbf{M}_{H}$ denotes the output of the hand decoder branch, which is supervised by $\textbf{G}_{h}$ provided in the dataset.

For inference, as stated in Sec. \ref{sec::ordm}, the decoder branches predict the masks of hands, new left-hand objects ${\textbf{M}}_{lo}$, and new right-hand objects ${\textbf{M}}_{ro}$.
Then the CODS provides the mask generation step to obtain the final masks of left-hand objects $\textbf{M}_O^L$, right-hand objects ${\textbf{M}}_O^R$, and two-hand objects ${\textbf{M}}_O^T$ through Equ. \ref{equ:codsrefer}-\ref{equ:codsreferend}.
The overall diagram for training and inference is shown in Algorithm \ref{algorithm}.

\section{Experiments}
\subsection{Datasets and Metrics}

To validate the effectiveness of the proposed CaRe-Ego, we compare its performance with different methods on the EgoHOS \citep{DBLP:conf/eccv/ZhangZSS22} in-domain test set. Additionally, we employed the EgoHOS out-of-domain test set as well as a dataset we developed from the HOI4D \citep{DBLP:conf/cvpr/LiuLJLWSLFWY22} dataset, referred to as mini-HOI4D, to assess the generalization capability of our CaRe-Ego.

\textbf{EgoHOS.}
The EgoHOS \citep{DBLP:conf/eccv/ZhangZSS22} dataset consists of 11,743 egocentric images containing per-pixel segmentation labels of hands and interacting objects gathered from Ego4D \citep{DBLP:conf/cvpr/GraumanWBCFGH0L22}, EPICKITCHEN \citep{DBLP:journals/corr/abs-1804-02748}, THU-READ \citep{DBLP:conf/icip/TangTLF017}, and their own collected egocentric videos. Among them, 8,993 images are used for training, 1,124 images are for validation, 1,126 images are used for in-domain testing, and 500 images are used for out-of-domain testing.

\textbf{mini-HOI4D.}
In order to comprehensively validate the generalization of the proposed method, we found another out-of-distribution (OOD) dataset - HOI4D \citep{DBLP:conf/cvpr/LiuLJLWSLFWY22} and further processed and aligned the included annotations to ensure its suitability for the EgoIHOS task requirements.
Ultimately, we generated a new dataset, named mini-HOI4D, comprising 1,095 egocentric images with corresponding hands and interacting object annotations, which can be used to assess and compare the generalization capabilities of our CaRe-Ego and other approaches.

This paper uses the most common and widely used metrics in the segmentation task to evaluate the performance, \emph{i.e.,} Intersection-over-Union (IoU) and pixel Accuracy (Acc) for each category. Mean IoU (mIoU) and mean accuracy (mAcc) are also employed to indicate the overall performance. 

\subsection{Implementation Details}
The experiments were conducted on 4 NVIDIA RTX 6000 Ada GPUs. The batch size was set to 12. Data preparation included cropping the image to (448, 448) and normalization with the mean of [106.011, 95.400, 87.429], and the standard variation was set to [64.357, 60.889, 61.419]. Following previous work \citep{DBLP:conf/eccv/ZhangZSS22}, we also used the Swin Transformer as the backbone. The patch size was set to 4, and the window size was set to 12. The AdamW optimizer was used in the experiment, the learning rate was set to 1e-5, and the weight decay was set to 0.01. 
The learning rate was set dynamically, \emph{i.e.,} from 0 to 10,000 iterations, the learning rate increased linearly to 1e-4, and from 10,000 iterations to the maximum iteration, the learning rate decreased linearly to 0. The maximum number of iterations in the experiment was set to 180,000.

\subsection{Quantitative Results}
\subsubsection{In-domain Results}

\begin{table*}[!ht]
\centering 
\Large	
\caption{Comparison results on the EgoHOS in-domain test set measured by IoU/Acc and mIoU/mAcc.}
\renewcommand\arraystretch{1.5}
\resizebox{\linewidth}{!}{
\begin{tabular}{lcccccccc} \hline 
\toprule
\multirow{2}{*}{Method} &\multirow{2}{*}{Type}  & \multirow{2}{*}{Backbone} & Left hand  & Right hand  & Left-hand objects & Right-hand objects  & Two-hand objects  & Overall 
\\ \cline{4-9}
& & & IoU(\%)$\, | \,$Acc(\%)
&IoU(\%)$\, | \,$Acc(\%)
&IoU(\%)$\, | \,$Acc(\%)
&IoU(\%)$\, | \,$Acc(\%)
&IoU(\%)$\, | \,$Acc(\%)
&mIoU(\%)$\, | \,$mAcc(\%)
\\ \midrule
$\text{Segformer}$\citep{DBLP:conf/nips/XieWYAAL21} 
&one-stage &MiT\citep{DBLP:conf/nips/XieWYAAL21}
&62.49$\, | \,$75.47
&64.77$\, | \,$78.13
&4.03$\, | \,$4.57
&3.01$\, | \,$3.17
&5.13$\, | \,$5.57
&27.89$\, | \,$33.38
\\ \hline
SCTNet\citep{DBLP:conf/aaai/XuWYCSG24} 
&one-stage
&SCTNet\citep{DBLP:conf/aaai/XuWYCSG24}
&81.94$\, | \,$90.25
&82.12$\, | \,$89.92
&17.77$\, | \,$24.49
&16.60$\, | \,$20.79
&21.74$\, | \,$29.08
&44.03$\, | \,$50.91
\\ \hline
${\text{Para.}^\prime}$\citep{DBLP:conf/eccv/ZhangZSS22} 
&one-stage
& Swin-B 
& 77.57$\, | \,$-
& 81.06$\, | \,$-
& 54.83$\, | \,$-
& 38.48$\, | \,$-
&39.14$\, | \,$-
& 58.22$\, | \,$-
\\ \hline
$\text{Segformer}$\citep{DBLP:conf/nips/XieWYAAL21} 
&one-stage  
&Swin-B
&65.36$\, | \,$78.20
&66.26$\, | \,$79.21
&11.63$\, | \,$15.45
&8.07$\, | \,$9.24
&12.48$\, | \,$15.97
&32.76$\, | \,$39.61
\\ \hline
$\text{Para.}$\citep{DBLP:conf/eccv/ZhangZSS22} 
& one-stage
& Swin-B 
& 69.08$\, | \,$75.57
& 73.50$\, | \,$75.93
& 48.67$\, | \,$39.50
& 36.21$\, | \,$39.33
&37.46$\, | \,$42.58
&52.98$\, | \,$54.58 
\\ \hline
$\text{Segmenter}$\citep{DBLP:conf/iccv/StrudelPLS21} 
&one-stage 
&Swin-B
&82.20$\, | \,$89.87
&83.28$\, | \,$91.92
&46.22$\, | \,$62.69
&34.79$\, | \,$45.59
&51.10$\, | \,$62.78
&59.52$\, | \,$70.57
\\ \hline
$\text{UperNet}$\citep{DBLP:conf/eccv/XiaoLZJS18} 
&one-stage 
&Swin-B 
&89.88$\, | \,$89.86
&91.39$\, | \,$91.32
&36.22$\, | \,$37.24
&40.55$\, | \,$42.26
&45.54$\, | \,$49.27
&60.71$\, | \,$61.99
\\ \hline 
Segmenter\citep{DBLP:conf/iccv/StrudelPLS21} &one-stage
&ViT\citep{DBLP:conf/iclr/DosovitskiyB0WZ21} &88.47$\, | \,$95.66 
&89.29$\, | \,$95.08
&49.87$\, | \,$61.14
&40.60$\, | \,$57.07
&46.96$\, | \,$60.81
&63.04$\, | \,$73.95
\\ \hline
MaskFormer\citep{DBLP:conf/nips/ChengSK21} &one-stage
&Swin-B
&90.45$\, | \,$95.90
&91.95$\, | \,$96.41
&43.51$\, | \,$67.08
&41.04$\, | \,$52.91
&54.65$\, | \,$64.86
&64.32$\, | \,$75.43
\\ \hline
Mask2Former\citep{DBLP:conf/cvpr/ChengMSKG22}
&one-stage
&Swin-B
&90.74$\, | \,$96.01
&92.25$\, | \,$96.20
&44.22$\, | \,$53.97
&46.05$\, | \,$58.10
&51.13$\, | \,$60.48
&64.88$\, | \,$72.95
\\ \hline \hline
\text{Seq.}\citep{DBLP:conf/eccv/ZhangZSS22}
&multi-stage 
& Swin-B 
&73.17$\, | \,$-
&80.56$\, | \,$-
&54.83$\, | \,$-
&38.48$\, | \,$-
&39.14$\, | \,$-
&57.24$\, | \,$-
\\ \hline
${\text{Seq.}}^\diamondsuit$\citep{DBLP:conf/eccv/ZhangZSS22} 
&multi-stage 
& Swin-B 
&77.25$\, | \,$-
&81.17$\, | \,$- 
&59.05$\, | \,$- 
&40.85$\, | \,$-
&49.94$\, | \,$-
&61.65$\, | \,$-
\\ \hline
${\text{Seq.}^\prime}$\citep{DBLP:conf/eccv/ZhangZSS22} 
&multi-stage 
& Swin-B 
&87.70$\, | \,$-
&88.79$\, | \,$-
&58.32$\, | \,$-
&40.18$\, | \,$-
&46.24$\, | \,$-
&64.25$\, | \,$- 
\\ \hline
${\text{Seq.}^\flat}$\citep{DBLP:conf/eccv/ZhangZSS22} 
&multi-stage 
& Swin-B 
&87.70$\, | \,$95.77
&88.79$\, | \,$91.29
&\textbf{\color{blue}62.20}$\, | \,$66.67
&44.40$\, | \,$59.85
&52.77$\, | \,$62.21
&67.17$\, | \,$75.16
\\ \hline \hline
\rowcolor{gray!40}
\textbf{CaRe-Ego} 
&one-stage 
& Swin-B 
&\color{blue}\textbf{92.34 $\, | \,$96.64}
&\color{blue}\textbf{93.64 $\, | \,$96.81}
&60.07 $\, | \,$\color{blue}\textbf{71.79}
&\color{blue}\textbf{56.69 $\, | \,$68.71}
&\color{blue}\textbf{54.73 $\, | \,$65.85} 
&\color{blue}\textbf{71.49 $\, | \,$79.96}\\
\bottomrule[1.2pt]
  \end{tabular}}
\label{tab::1compare}
\end{table*}

\emph{Comparison results on EgoHOS in-domain test set.}
To validate the effectiveness of the proposed CaRe-Ego, we conducted experiments by training on the EgoHOS train set and testing on the EgoHOS in-domain test set. 
To ensure a comprehensive comparison, we incorporate all existing one-stage and multi-stage methods \citep{DBLP:conf/eccv/ZhangZSS22,DBLP:conf/nips/XieWYAAL21,DBLP:conf/cvpr/Jain0C0OS23,DBLP:conf/eccv/CaoWCJZTW22,DBLP:conf/eccv/XiaoLZJS18,DBLP:conf/cvpr/ChengMSKG22,DBLP:conf/iccv/StrudelPLS21} tailored for the EgoIHOS task in this evaluation. The results are presented in Table \ref{tab::1compare}.
In general, the proposed CaRe-Ego far outperforms all previous methods in most categories on both metrics.
In particular, compared with the previous best one-stage method - Mask2Former \citep{DBLP:conf/cvpr/ChengMSKG22}, our method has demonstrated pronounced advancements across various categories and exhibited particularly notable enhancements in object segmentation. For example, our method showcases substantial enhancements of 15.85\%, 10.64\%, and 3.6\% on left-, right-, and two-hand objects, respectively. 
Furthermore, compared with the results of the previous best multi-stage methods $\text{Seq}^\flat$, our CaRe-Ego improves IoU by more than 4\% on both the left and right hands. The most substantial enhancement is observed in the right-hand objects, surpassing 12\%. 
Finally, compared with the previous best method, our proposed method attains a commendable advancement of 4.33\% and 4.55\% in mIoU and mAcc, respectively.
These results demonstrate the outstanding performance achieved by the proposed CaRe-Ego, especially on the interacting object segmentation.
This achievement is mainly because 1) the proposed HOFE explicitly models the hand-object interactive relationship to enhance the extracted object features, and 2) the designed CODS emphasizes interaction feature learning through decoupling the correlations between diverse object categories. 

\begin{table*}[!ht]
\centering 
\Large	
\caption{Comparison results on the EgoHOS out-of-domain test set.}
\renewcommand\arraystretch{1.25}
\resizebox{\linewidth}{!}{
\begin{tabular}{lcccccccc} \hline 
\toprule
\multirow{2}{*}{Method} &\multirow{2}{*}{Type}  & \multirow{2}{*}{Backbone} & Left hand  & Right hand  & Left-hand objects & Right-hand objects  & Two-hand objects  & Overall 
\\ \cline{4-9}
& & & IoU(\%)$\, | \,$Acc(\%)
&IoU(\%)$\, | \,$Acc(\%)
&IoU(\%)$\, | \,$Acc(\%)
&IoU(\%)$\, | \,$Acc(\%)
&IoU(\%)$\, | \,$Acc(\%)
&mIoU(\%)$\, | \,$mAcc(\%)
\\ \midrule
$\text{Segformer}$\cite{DBLP:conf/nips/XieWYAAL21} 
&one-stage &MiT\cite{DBLP:conf/nips/XieWYAAL21}
&71.97$\, | \,$87.33
&71.44$\, | \,$81.75
&7.60$\, | \,$8.34
&5.00$\, | \,$5.65
&4.91$\, | \,$5.95
&32.18$\, | \,$37.80
\\ \hline
$\text{Segformer}$\cite{DBLP:conf/nips/XieWYAAL21} 
&one-stage  
&Swin-B
&74.01$\, | \,$86.80
&70.84$\, | \,$79.44
&15.61$\, | \,$21.77
&7.32$\, | \,$9.56
&8.29$\, | \,$11.67
&35.21$\, | \,$41.85
\\ \hline
SCTNet\cite{DBLP:conf/aaai/XuWYCSG24} 
&one-stage
&SCTNet\cite{DBLP:conf/aaai/XuWYCSG24}
&87.12$\, | \,$94.62
&86.29$\, | \,$90.92
&31.18$\, | \,$49.14
&19.70$\, | \,$27.47
&13.32$\, | \,$17.12
&47.52$\, | \,$55.85
\\ \hline
${\text{UperNet}}$\cite{DBLP:conf/eccv/XiaoLZJS18} 
&one-stage 
& Swin-B 
&93.17$\, | \,$96.89	
&93.96$\, | \,$96.00
&42.53$\, | \,$64.83
&28.88$\, | \,$54.59	
&24.35$\, | \,$27.83
&56.58$\, | \,$68.03		
\\ \hline 
$\text{Segmenter}$\cite{DBLP:conf/iccv/StrudelPLS21} 
&one-stage 
&Swin-B
&83.13$\, | \,$91.22
&84.85$\, | \,$92.56
&{\textbf{\color{blue}57.97}}$\, | \,$72.44
&38.59$\, | \,$52.75
&\textbf{\color{blue}44.98}$\, | \,$52.88
&61.90$\, | \,$72.37
\\ \hline
${\text{Maskformer}}$\cite{DBLP:conf/nips/ChengSK21} 
&one-stage 
& Swin-B 
&92.69$\, | \,$95.58	
&94.02$\, | \,$96.10
&51.81$\, | \,$70.53
&39.84$\, | \,$60.49	
&39.43$\, | \,$46.52
&63.56$\, | \,$73.84
\\ \hline
${\text{Mask2former}}$\cite{DBLP:conf/cvpr/ChengMSKG22} 
&one-stage 
& Swin-B 
&91.46$\, | \,$97.05	
&93.04$\, | \,$96.38
&53.41$\, | \,$64.39
&\textbf{\color{blue}44.90$\, | \,$64.18}	
&35.61$\, | \,$39.78
&63.68$\, | \,$72.36
\\ \hline				
Segmenter\cite{DBLP:conf/iccv/StrudelPLS21} &one-stage
&ViT\cite{DBLP:conf/iclr/DosovitskiyB0WZ21} 
&89.40$\, | \,$95.02
&90.58$\, | \,$94.86
&52.73$\, | \,$\textbf{\color{blue}75.75}
&43.88$\, | \,$56.34
&42.33$\, | \,$51.08
&63.78$\, | \,$74.61
\\ \hline \hline
${\text{Seq.}^\flat}$\cite{DBLP:conf/eccv/ZhangZSS22} 
&multi-stage 
& Swin-B 
&81.77$\, | \,$87.83
&78.82$\, | \,$85.98
&46.93$\, | \,$57.17
&26.40$\, | \,$43.85
&42.38$\, | \,$\textbf{\color{blue}54.76}
&55.26$\, | \,$65.92
\\ \hline \hline
\rowcolor{gray!40}
\textbf{CaRe-Ego} 
&one-stage 
& Swin-B 
&\color{blue}\textbf{94.47 $\, | \,$97.09}
&\color{blue}\textbf{94.41 $\, | \,$96.69}
&51.56 $\, | \,$72.30
&36.80 $\, | \,$60.90
&41.84 $\, | \,$46.28
&\color{blue}\textbf{63.82 $\, | \,$74.65}\\
\bottomrule[1.2pt]
  \end{tabular}}
\label{tab::2compare}

\end{table*}

\subsubsection{Out-of-domain Results}
\emph{1) Comparison results on EgoHOS out-of-domain test set.}
In order to test the generalization performance of different models, the EgoHOS dataset provides 500 out-of-domain images, which are captured from YouTube and annotated manually.
Thus, We compare CaRe-Ego with different one-stage and multi-stage models \citep{DBLP:conf/eccv/ZhangZSS22,DBLP:conf/nips/XieWYAAL21,DBLP:conf/cvpr/Jain0C0OS23,DBLP:conf/eccv/CaoWCJZTW22,DBLP:conf/eccv/XiaoLZJS18,DBLP:conf/cvpr/ChengMSKG22,DBLP:conf/iccv/StrudelPLS21} by directly testing saved best models on the EgoHOS out-of-domain test set, results are shown in Table \ref{tab::2compare}. 
It can be observed that the results of our proposed CaRe-Ego on the two metrics of most categories have obtained significant generalization performance.
Among them, the IoU of the left and right hand exceed 94\%, and the Acc. of these two categories exceed 96\%. 
The segmentation results on various object categories also achieve balanced performance. Finally, our CaRe-Ego achieves state-of-the-art performance in overall mIoU and mAcc metrics and demonstrates excellent generalization performance on the out-of-domain test set. 
These advantages are primarily due to our proposed CaRe-Ego emphasizing the contact learning between hands and objects through HOFE and CODS regardless of the background change.

\begin{table*}[!ht]
\centering 
\Large	
\caption{Comparison results on the mini-HOI4D test set.}
\renewcommand\arraystretch{1.5}
\resizebox{\linewidth}{!}{
\begin{tabular}{lccccccc} \hline 
\toprule
\multirow{2}{*}{Method} &\multirow{2}{*}{Type}  & \multirow{2}{*}{Backbone} & Left hand  & Right hand   & Right-hand objects  & Two-hand objects  & Overall 
\\ \cline{4-8}
& & & IoU(\%)$\, | \,$Acc(\%)
&IoU(\%)$\, | \,$Acc(\%)
&IoU(\%)$\, | \,$Acc(\%)
&IoU(\%)$\, | \,$Acc(\%)
&mIoU(\%)$\, | \,$mAcc(\%)
\\ \midrule
$\text{Segformer}$\cite{DBLP:conf/nips/XieWYAAL21} 
&one-stage  
&Swin-B
&27.73$\, | \,$87.97
&53.27$\, | \,$66.91
&8.43$\, | \,$9.72
&9.30$\, | \,$13.40
&24.68$\, | \,$44.50
\\ \hline
$\text{Segformer}$\cite{DBLP:conf/nips/XieWYAAL21} 
&one-stage &MiT\cite{DBLP:conf/nips/XieWYAAL21}
&30.16$\, | \,$92.13
&56.44$\, | \,$72.42
&5.17$\, | \,$5.52
&12.02$\, | \,$13.41
&25.95$\, | \,$45.87
\\ \hline
SCTNet\cite{DBLP:conf/aaai/XuWYCSG24} 
&one-stage
&SCTNet\cite{DBLP:conf/aaai/XuWYCSG24}
&35.83$\, | \,$95.25
&66.29$\, | \,$71.27
&17.72$\, | \,$22.68
&20.98$\, | \,$29.98
&35.21$\, | \,$54.80
\\ \hline
$\text{Segmenter}$\cite{DBLP:conf/iccv/StrudelPLS21} 
&one-stage 
&Swin-B
&70.31$\, | \,$89.02
&73.78$\, | \,$89.23
&21.82$\, | \,$43.75
&44.98$\, | \,$56.97
&52.72$\, | \,$69.74
\\ \hline
${\text{UperNet}}$\cite{DBLP:conf/eccv/XiaoLZJS18} 
&one-stage 
& Swin-B 
&54.82$\, | \,$97.71	
&84.43$\, | \,$86.04
&20.34$\, | \,$25.77	
&29.34$\, | \,$36.58
&47.23$\, | \,$61.53	
\\ \hline
${\text{MaskFormer}}$\cite{DBLP:conf/nips/ChengSK21} 
&one-stage 
& Swin-B 
&58.50$\, | \,$96.63
&83.66$\, | \,$87.83
&\textbf{\color{blue}35.28}$\, | \,$44.81
&56.91$\, | \,$73.47
&58.59$\, | \,$75.69
\\ \hline
Segmenter\cite{DBLP:conf/iccv/StrudelPLS21} &one-stage
&ViT\cite{DBLP:conf/iclr/DosovitskiyB0WZ21} &\textbf{\color{blue}74.70}$\, | \,$94.03
&85.58$\, | \,$92.34
&22.38$\, | \,$44.42
&58.67$\, | \,$73.99
&60.33$\, | \,$76.20
\\ \hline
${\text{Mask2Former}}$\cite{DBLP:conf/cvpr/ChengMSKG22} 
&one-stage 
& Swin-B 
&70.13$\, | \,$97.48
&88.57$\, | \,$89.38
&32.37$\, | \,$45.22
&55.72$\, | \,$\textbf{\color{blue}74.17}
&61.70$\, | \,$76.56
\\ \hline \hline
${\text{Seq.}^\flat}$\cite{DBLP:conf/eccv/ZhangZSS22} 
&multi-stage 
& Swin-B 
&8.74$\, | \,$40.90
&34.60$\, | \,$38.05
&23.88$\, | \,$28.99
&53.96$\, | \,$61.67
&30.30$\, | \,$42.40
\\ \hline \hline
\rowcolor{gray!40}
\textbf{CaRe-Ego} 
&one-stage 
& Swin-B 
&70.39 $\, | \,$\color{blue}\textbf{97.79}
&\color{blue}\textbf{89.76 $\, | \,$93.09}
&27.56 $\, | \,$\color{blue}\textbf{49.35}
&\textbf{\color{blue}60.08} $\, | \,$68.01 
&\color{blue}\textbf{61.95 $\, | \,$77.06}\\
\bottomrule[1.2pt]
  \end{tabular}}
\label{tab::3compare}
\end{table*}

\emph{2) Comparison results on mini-HOI4D dataset.}
In order to comprehensively validate the generalizability of the proposed method, we also assess the generalization capabilities of algorithms using a completely different mini-HOI4D dataset.
It should be noted that since the original HOI4D \cite{DBLP:conf/cvpr/LiuLJLWSLFWY22} dataset contains very few samples of left-hand objects, we do not perform testing on left-hand object category.
We also utilized the saved best models to test on the mini-HOI4D dataset without further training procedures, the comparison results are shown in Table \ref{tab::3compare}.
Overall, our proposed CaRe-Ego outperforms all the previous one-stage and multi-stage methods \citep{DBLP:conf/eccv/ZhangZSS22,DBLP:conf/nips/XieWYAAL21,DBLP:conf/cvpr/Jain0C0OS23,DBLP:conf/eccv/CaoWCJZTW22,DBLP:conf/eccv/XiaoLZJS18,DBLP:conf/cvpr/ChengMSKG22,DBLP:conf/iccv/StrudelPLS21} concerning mIoU and mAcc. 
Our method excels notably in hand segmentation, \emph{i.e.,} 1.57\% and 3.71\% improvements on IoU and Acc for the right hand compared with the previous best Mask2Former model.
Our method also competes well in interacting object segmentation, yielding commendable outcomes.
Therefore, it can be deduced from the results of the EgoHOS out-of-domain and mini-HOI4D test sets that our proposed CaRe-Ego exhibits superior generalization capabilities in contrast to the comparative methods, which benefits from the designed HOFE and CODS aiming to establish the correlation between \emph{hands and objects} as well as  \emph{objects and objects}, thereby improving the feature learning and segmentation ability of the network.

\begin{table}[!t]
\scriptsize
\centering
\caption{Ablation results on the EgoHOS in-domain test set under the same setting measured by IoU and mIoU. Lh: left hand, rh: right hand, lo: left-hand object, ro: right-hand object, to: two-hand object.}
\begin{tabular}{lcccccc}
\toprule[1.4pt]
Method   &Lh (\%) & Rh (\%) &Lo (\%) & Ro (\%) &To (\%) &mIoU (\%) \\ 
\midrule[1pt]
Basic &90.47 &91.02 &44.11&45.57&51.08& 64.45 \\ \hline
Basic+CODS  &\textbf{93.14}&\textbf{93.73}&\textbf{60.86}&54.58&49.69 &70.40 \\ \hline
Basic+HOFE &92.76&93.36&44.15&53.68&51.85&67.16 \\ \hline
Basic+CODS+HOFE  &92.34&93.64&60.07&\textbf{56.69}&\textbf{54.73}&\textbf{71.49} \\ 
\bottomrule[1.2pt]
\end{tabular}
\label{tab::abla}
\end{table}

\subsubsection{Ablation Study}
In order to verify the effectiveness of the proposed method alongside its corresponding sub-module and strategy, we conduct the ablation study in this section.
The experimental results are shown in Table \ref{tab::abla}. 
The basic model uses Swin-B as the encoder and uses three decoder branches to predict hands, contact boundaries, and objects.
This basic model is a naive solution for the EgoIHOS task and does not model any relationships between hands and interacting objects, so its results are relatively poor.
The $2^{nd}$ and $3^{rd}$ rows of the table show the results after adding the CODS and HOFE independently, which proves that adding them to the basic model is beneficial.  
 The last row is the result after integrating the CODS and the HOFE. By explicitly modeling the relations between \emph{hands and objects} as well as \emph{objects and objects}, the segmentation accuracy achieves its peak performance.
 Generally speaking, using HOFE and CODS simultaneously can further improve the segmentation performance, which verifies the effectiveness of our proposed method and sub-methods.

 \begin{table}[!t]
\scriptsize
\centering
\caption{The experimental results using different hyper-parameters. $\alpha$,  $\gamma$, and $\lambda$ are the weights of objects, hands, and CB losses.}
{
\begin{tabular}{c|c|c|c|c|c|c|c|c}
\bottomrule[1.25pt]
$\alpha$ &$\gamma$ &$\lambda$ &Left hand (\%) & Right hand (\%) &Left-hand object (\%) & Right-hand object (\%) &Two-hand object (\%) &mIoU (\%)  \\ \hline
1.0&1.0&1.0&90.66&91.80&57.38&56.20&52.12&69.63 \\ \hline
1.0&0.5&0.5&89.74&90.56&57.8&55.46&51.08&68.93 \\ \hline
0.5&0.5&1.0&90.22&91.10&57.78&55.59&50.81&69.10 \\ \hline
0.5&1.0&0.5&\textbf{92.34}&\textbf{93.64}&\textbf{60.07}&\textbf{56.69}&\textbf{54.73}&\textbf{71.49} \\
\toprule[1.25pt]
\end{tabular}}
\label{tab::hyper}
\end{table}
 
\subsubsection{Hyper-parameter Study}
The method presented in this paper requires setting multiple hyper-parameters for total training loss, \emph{i.e.,} $\alpha, \gamma$, and $\lambda$. 
This section conducts experiments on the configurations of different hyper-parameters, results are shown in Table \ref{tab::hyper}. We assign the values of $\alpha$, $\gamma$, and $\lambda$ based on the relative significance of the interacting objects, hands, and contact boundaries. 
In the $1^{st}$ row, all categories are considered equal importance, while in subsequent experiments (\emph{i.e.,} $2^{nd}, 3^{rd}$ and $4^{th}$), we regard the objects, CBs, and hands as the most crucial, respectively.
The results lead to the inference that amplifying the importance of hands can yield enhancements in the method's performance, which can be explained intuitively because the proposed HOFE leverages hands as prior knowledge for object feature enhancement.

\begin{figure*}[!t]
\centering
\includegraphics[width=\linewidth]{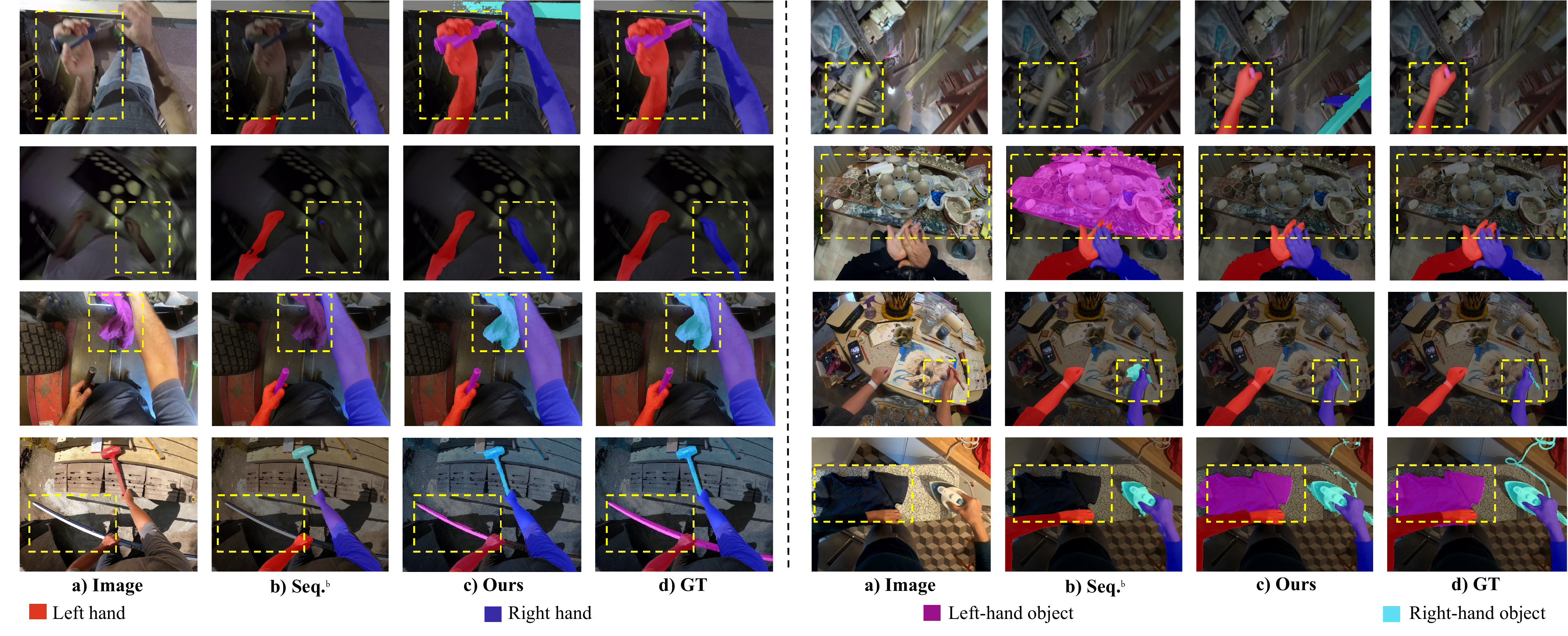}%
\caption{Visualization results of the CaRe-Ego compared with the multi-stage method $\text{Seq.}^\flat$ on the EgoIHOS in-domain test set. The main improvements are highlighted in the dashed yellow box.}
\label{fig_4}
\end{figure*}

\begin{figure}[!t]
\centering
\includegraphics[scale=0.25]{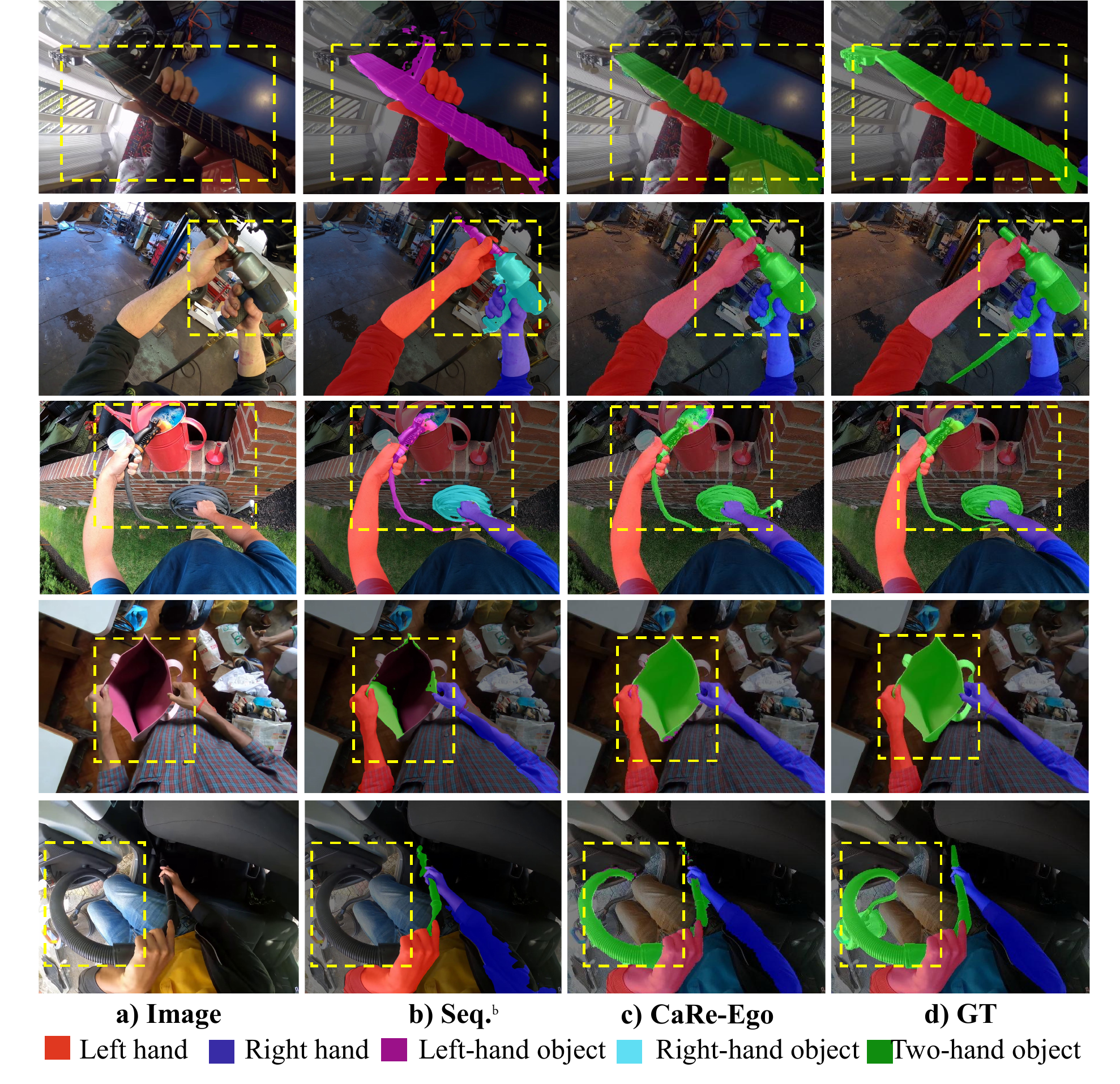}%
\caption{Visualization results of \textbf{two-hand objects} on the EgoHOS in-domain test set compared with multi-stage method $\text{Seq}^\flat$. The main improvements are highlighted in the dashed yellow box.}
\label{fig_5}
\end{figure}

\subsection{Qualitative Results}
\subsubsection{In-domain Results}

\emph{1) Results on EgoHOS in-domain test set.} This section shows the visualization results on the EgoHOS in-domain test set compared with the previous best multi-stage method $\text{Seq.}^\flat$, as shown in Fig. \ref{fig_4}. 
We can observe that our method has better performance in segmenting hands, as shown in the results of the $1^{st}$ and $2^{nd}$ rows. Moreover, since our method proposes an CODS to model the relationship between diverse objects, CaRe-Ego also has better consistency and accuracy when segmenting interacting objects, as shown in the results of the $3^{rd}$ and $4^{th}$ rows.
Furthermore, although our method does not predict two-hand objects through the learnable network, the performance is also more significant than the compared method, which is shown in Fig. \ref{fig_5}. 
In the results of the comparative method, the two-hand objects or part of them may be segmented into left- and right-hand objects. However, our method only focuses on the objects that the hands interact with during the learning process using the CODS rather than concentrating on ``whether the object interacts with the left and right hands at the same time," thereby enhancing the segmentation consistency and accuracy of two-hand objects. 

\begin{figure*}[!t]
\centering
\includegraphics[width=\linewidth]{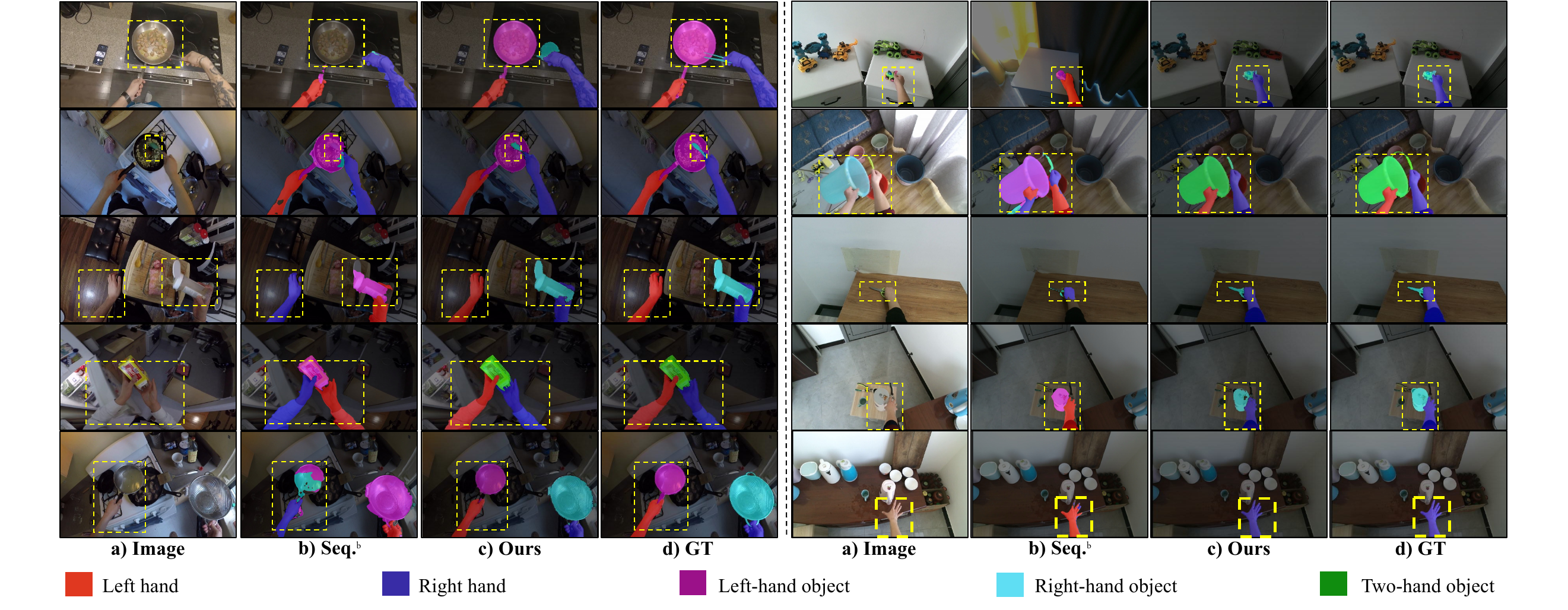}%
\caption{Visualization results of the CaRe-Ego compared with the multi-stage method $\text{Seq.}^\flat$ on the EgoHOS out-of-domain test set (left) and mini-HOI4D dataset (right).}
\label{fig_6}
\end{figure*}

\subsubsection{Out-of-domain Results}

\emph{1) Results on EgoHOS out-of-domain test set.}
We visualize the comparative results on the EgoHOS out-of-domain test set with the $\text{Seq}^\flat$ method in Fig. \ref{fig_6} (left). The figure shows that our CaRe-Ego has stronger generalization than the compared multi-stage model. Specifically, our method can segment and identify left and right hands accurately, as shown in the $3^{rd}$ to the last rows of results. Besides, our method can precisely segment and recognize different kinds of objects, ensuring consistent segmentation of interacting objects.

\emph{2) Results on mini-HOI4D dataset.}
We visualize the comparative results on the mini-HOI4D test set with the $\text{Seq}^\flat$ method in Fig. \ref{fig_6} (right).
It can be observed that the contrasting multi-stage method easily confuses the left and right hands, such as $1^{st}$, $2^{nd}$, $4^{th}$, and $5^{th}$ rows of the results. 
In contrast, our method exhibits robust generalization capabilities coupled with outstanding segmentation performance.
These visualization results on out-of-distribution datasets exhibit the superior generalization performance of our proposed CaRe-Ego, which is primarily due to the designed HOFE and CODS can establish the correlation between \emph{hands and objects} as well as \emph{objects and objects}, thereby emphasizing the hand-object interaction learning and improving the segmentation ability of the network.

\section{Conclusion}
This paper introduces the CaRe-Ego to effectively handle the EgoIHOS task while achieving high segmentation accuracy by incorporating two parts to emphasize the interaction learning. First, the HOFE is proposed to facilitate the modeling of hand-object interactions through hand-guided cross-attention, which can utilize hand features as prior knowledge to extract more contact-relevant and representative object features.
Second, we design an ingenious CODS to decouple the relationships among diverse interacting object categories, thereby emphasizing interaction learning and reducing the confusion of the network about object classification.
CaRe-Ego achieves
state-of-the-art performance on the EgoHOS in-domain test set and two out-of-distribution test sets. Comprehensive ablation studies demonstrate the effectiveness of HOFE and CODS in our architecture.

\textbf{Limitations and Future Work}
The EgoIHOS task inherently possesses certain limitations, as it is only capable of predicting the hand that interacts with an object, without the ability to infer the object's name or semantic information. To address this issue, we consider enhancing and refining the EgoIHOS task in future work. Specifically, we aim to incorporate the prediction of object semantic information alongside the hand-object segmentation, thereby expanding the practical applicability of the algorithm in real-world scenarios.

\section{Acknowledgements}
The research work was conducted in the JC STEM Lab of Machine Learning and Computer Vision funded by The Hong Kong Jockey Club Charities Trust.


\printcredits

\bibliographystyle{cas-model2-names}

\bibliography{cas-refs}


\end{document}